\def\isarxiv{}

\ifdefined\isarxiv
    \documentclass[sigconf,nonacm]{acmart}
    \pdfoutput=1
\else
    \documentclass[sigconf,review,anonymous]{acmart}
\fi

\AtBeginDocument{%
  \providecommand\BibTeX{{%
    \normalfont B\kern-0.5em{\scshape i\kern-0.25em b}\kern-0.8em\TeX}}}

\setcopyright{acmlicensed}
\copyrightyear{2026}
\acmYear{2026}
\acmDOI{XXXXXXX.XXXXXXX}

\acmConference[HSCC/ICCPS ’26]{29th ACM International Conference on Hybrid Systems: Computation and Control / 17th ACM/IEEE International Conference on Cyber-Physical Systems (CPS-IoT Week 2026)}{May 11–14, 2026}{Saint-Malo, France}
\acmISBN{978-1-4503-XXXX-X/26/05}

\usepackage{subcaption}
\usepackage{enumitem}

\usepackage{amsfonts, bm, mathtools, amsmath}

\def\Prob{{\mathbb{P}}}
\def\ProbQ{{\mathbb{Q}}}
\def\discProb{{\mathbb{D}}}
\def\sProb{{\mathcal{P}}}
\def\sDiscProb{{\mathcal{D}}}
\def\sigmaAlgebra{{\mathcal{B}}}

\def\signature{{\Delta}}

\def\GMM{\mathcal{G}}

\def\Ndist{\mathcal{N}}

\def\wasserstein{{\mathbb{W}}}

\def\indicator{{\mathcal{I}}}

\def\diag{\mathrm{diag}}

\newcommand{\elem}[2]{{#1^{{(#2)}}}}

\def\eucl{{\mathbb{R}}}
\def\realNum{{\mathbb{R}}}
\def\natNum{\mathbb{N}}

\def\sSimplex{\Pi}

\def\sA{{\mathcal{A}}}

\def\sC{{\mathcal{C}}}

\def\sP{{\mathcal{P}}}

\def\sR{{\mathcal{R}}}
\def\sS{{\mathcal{S}}}

\def\sX{{\mathcal{X}}}
\def\sY{{\mathcal{Y}}}

\def\bsR{{\bm{\mathcal{R}}}}

\def\mT{{{T}}}

\def\mV{{{V}}}

\def\vlambda{{{\lambda}}}
\def\vmu{{{\mu}}}
\def\vpi{{{\pi}}}

\def\va{{{a}}}
\def\vb{{{b}}}
\def\vc{{{c}}}

\def\vx{{{x}}}

\newcommand{\evlambda}[1]{{\vlambda^{(#1)}}}

\newcommand{\evpi}[1]{{\vpi^{(#1)}}}

\newcommand{\eva}[1]{{\va^{(#1)}}}
\newcommand{\evb}[1]{{\vb^{(#1)}}}

\newcommand{\evx}[1]{{\vx^{(#1)}}}

\usepackage{xcolor}

\usepackage{algorithm}
\usepackage{algpseudocode}
\usepackage{booktabs}
\usepackage{makecell}
\usepackage{amsmath,mathtools}

\usepackage{listings}
\usepackage[T1]{fontenc} 

\begin{document}

\title{\texttt{discretize\_distributions}: Efficient Quantization of Gaussian Mixtures with Guarantees in Wasserstein Distance}

\ifdefined\isarxiv
    \input{arxiv_title_layout}
\else
\fi

\author{Steven Adams}
\email{s.j.l.adams@tudelft.nl}
\affiliation{%
  \institution{TU Delft}
  \city{Delft}
  \country{the Netherlands}
}

\author{Elize Alwash}
\affiliation{%
  \institution{EPFL}
  \city{Lausanne}
  \country{Switzerland}
}

\author{Luca Laurenti}
\affiliation{%
  \institution{TU Delft}
  \city{Delft}
  \country{the Netherlands}
}

\begin{abstract}
We present \texttt{discretize\_distributions}, a Python package that efficiently constructs discrete approximations of Gaussian mixture distributions and provides guarantees on the approximation error in Wasserstein distance.
The package implements state-of-the-art quantization methods for Gaussian mixture models and extends them to improve scalability.
It further integrates complementary quantization strategies such as sigma-point methods and provides a modular interface that supports custom schemes and integration into control and verification pipelines for cyber-physical systems.
We benchmark the package on various examples, including high-dimensional, large, and degenerate Gaussian mixtures, and demonstrate that \texttt{discretize\_distributions} produces accurate approximations at low computational cost.
\end{abstract}


\maketitle

\section{Introduction}
Modern cyber-physical systems increasingly rely on probabilistic models to represent uncertainty from system dynamics and data-driven approximations \cite{stengel1994optimal, ljung2010perspectives}. 
Evaluating such models often requires propagating distributions through nonlinear transformations, operations that are rarely analytically tractable.
A common approach is therefore to approximate continuous distributions with discrete ones that can be efficiently propagated and integrated.
This process, known as quantization, has long been studied in information theory, where it underpins signal compression \cite{bucklew2003multidimensional, gersho2012vector}, and in numerical analysis, where it enables efficient integration w.r.t. continuous measures \cite{pages2004optimal}.
In the context of dynamical systems, quantization enables representing uncertainty with a finite set of points that can be efficiently propagated through the system dynamics. 
For many safety-critical applications, where correctness must be guaranteed, it is essential that such approximations come with formal accuracy guarantees.
This is particularly relevant for Gaussian mixtures, which are widely used for their ability to represent complex, multimodal behaviors while retaining analytical tractability \cite{bishop2006pattern}.
However, despite their widespread use, no efficient implementation exists for quantizing Gaussian mixtures with formal error guarantees, a gap addressed by \texttt{discretize\_distributions}, the tool introduced in this paper.

Quantization is a well-established concept in probability and statistics \cite{graf2000foundations} 
and is commonly divided into sample-based and constructive methods. 
Sample-based approaches rely on empirical distributions from random samples \cite{dereich2013constructive}, whereas constructive ones explicitly build a discrete approximation minimizing a distance metric, typically the Wasserstein distance.
The latter class is particularly attractive for model-based applications, as it enables computationally efficient and analytically tractable discretizations, although constructing optimal quantizations is highly nontrivial.
Optimization-based formulations such as \cite{pages2003optimal} employ stochastic gradient methods to construct quantizations but suffer from poor scalability. 
Heuristic constructions, including symmetric placements along principal axes \cite{hanebeck2009dirac},
are more efficient but lack formal guarantees. 
Moreover, both approaches generally require costly numerical integration for error estimation and are typically restricted to Gaussian distributions.
Recently, \cite{adams2024finite} proposed a highly efficient constructive approach to generate quantizations of Gaussian mixture distributions using precomputed optimal quantization templates. 
This approach naturally comes with formal convergence guarantees on the quantization error and has proven highly effective in practice \cite{figueiredo2025efficient, adams2025formal}.

In this work, we present \texttt{discretize\_distributions}, a Python package for efficiently constructing discrete approximations of Gaussian mixture distributions with formal error guarantees. Building on the framework of~\cite{adams2024finite}, it extends the method with a mode-wise quantization procedure that groups mixture components by their dominant modes, reducing redundancy and computational cost compared to component-wise schemes while preserving accuracy. The package supports closed-form evaluation of Wasserstein errors, includes alternative strategies such as sigma-point methods~\cite{hanebeck2009dirac}, and provides a modular interface for custom scheme construction and integration into broader probabilistic modeling, uncertainty-propagation, and verification pipelines.
In doing so, it turns previously theoretical quantization methods with guarantees into a practical, ready-to-use software component.

In what follows, Section~\ref{sec:theory} introduces Gaussian mixture distributions and the quantization operation, Section~\ref{sec:quantization_gmms} presents the algorithmic framework underlying the tool, Section~\ref{sec:tool} describes the package design and usage, and Section~\ref{sec:experiments} reports numerical benchmarks demonstrating accuracy, efficiency, and scalability.

\newpage
\section{Notation}
For a vector $\vx\in\eucl^n$, we denote by $\|\vx\|$ its Euclidean norm and by $\evx{i}$ its $i$-th element. 
For a region \(\sX\subset\realNum^n\), the indicator function \(\indicator_{\sX}(x)\) equals $1$ if \(x \in \sX\) and \(0\) otherwise. 
We write \(\sProb(\sX)\) for the set of probability distributions on \((\sX, \sigmaAlgebra(\sX))\), where \(\sigmaAlgebra(\sX)\) is the Borel \(\sigma\)-algebra, and \(\sProb_2(\sX)\) for the subset with finite second moments, i.e., those \(\Prob \in \sProb(\sX)\) satisfying \(\int_{\sX} \|\vx\|^2 \Prob(d\vx) < \infty\).
For $N \in \natNum$, $\sSimplex_N = \{\pi \in \eucl^N_{\ge 0} : \sum_{i=1}^N \evpi{i} = 1\}$ is the $N$-simplex.
A discrete probability distribution $\discProb \in \sProb(\sX)$ is defined as $\discProb = \sum_{i=1}^N \evpi{i}\delta_{\vc_i}$, where $\delta_{\vc}$ is the Dirac delta function centered at $\vc \in \sX$ and $\pi \in \sSimplex_N$. 
The set of discrete probability distributions on $\sX$ with at most $N$ locations is denoted by $\sDiscProb_N(\sX) \subset \sProb(\sX)$.
Finally, for measurable spaces \((\sX, \sigmaAlgebra(\sX))\) and \((\sY, \sigmaAlgebra(\sY))\), a measure \(\Prob \in \sProb(\sX)\), and a measurable mapping \(h : \sX \to \sY\), we denote by \(h\#\Prob\) the pushforward of \(\Prob\) under \(h\), defined as \((h\#\Prob)(\sA) = \Prob(h^{-1}(\sA))\) for all \(\sA \in \sigmaAlgebra(\sY)\).

\begin{figure}[h]
    \centering
    \includegraphics[width=0.8\linewidth]{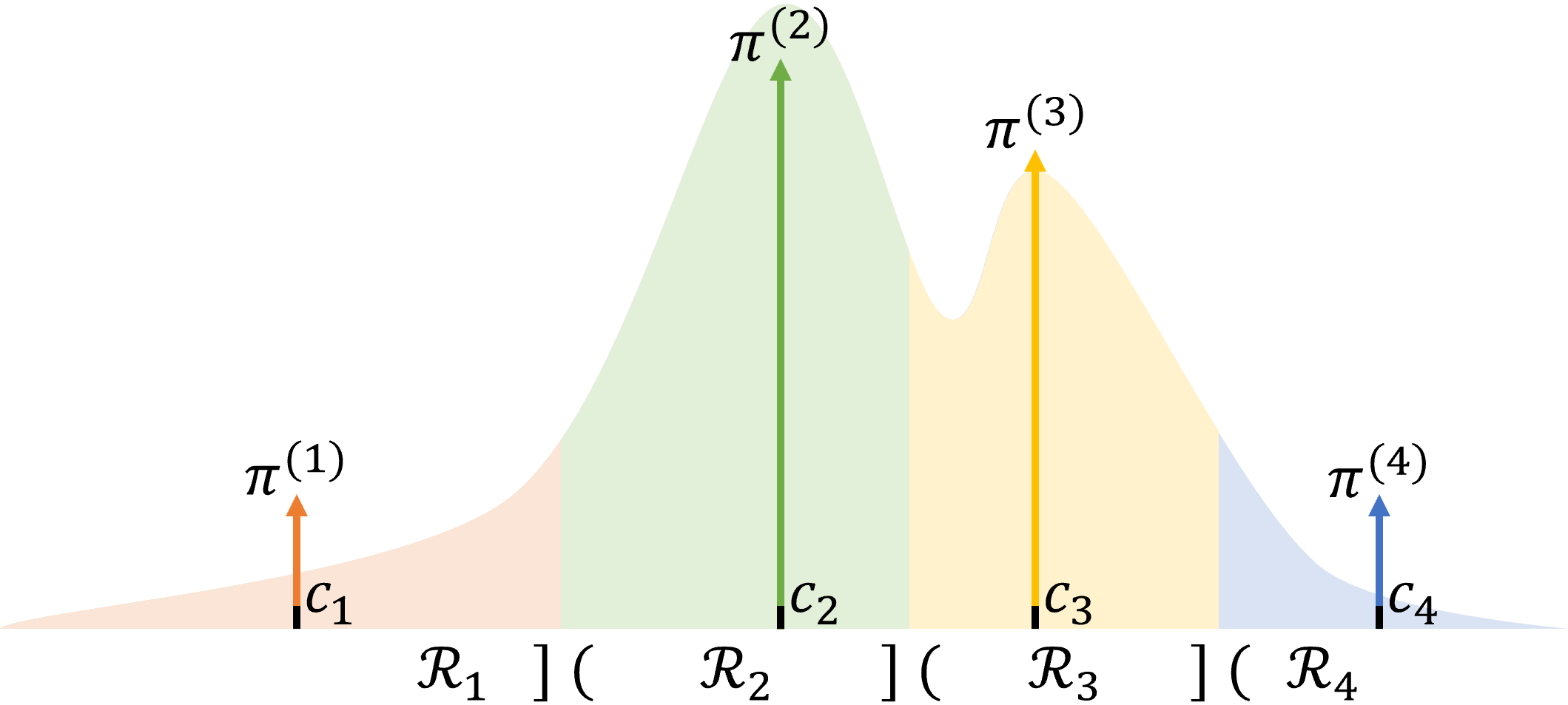}
    \caption{
    Illustration of the quantization of a continuous probability distribution $\Prob$, yielding its discrete approximation $\signature_{\bsR,\sC}\#\Prob$ with a support of size $4$. 
    The probability mass $\evpi{i}$ of \(\Prob\) in each region \(\sR_i\) is mapped to the location $\vc_i$.
    }
    \label{fig:quantization_operator}
\end{figure}

\section{Theoretical Background}\label{sec:theory}
This section summarizes the theoretical foundations underlying the tool, namely Gaussian mixture models, their quantization into discrete distributions, and the computation of the resulting approximation error in Wasserstein distance.

\subsection{Gaussian Mixture Distributions}
A probability distribution \( \Prob \in \sProb(\eucl^d) \) is called a \emph{Gaussian mixture distribution} with \( M \in \natNum \) components if
\(
    \Prob = \sum_{i=1}^M \elem{\vpi}{i} \Ndist(\vmu_i, \Sigma_i),
\)
where \( \vpi \in \Pi_M \) denotes the mixture weights, and \( \vmu_i \in \eucl^d \) and \( \Sigma_i \in \eucl^{d \times d} \) are the mean and covariance matrix of the \( i \)-th Gaussian component, respectively.
Gaussian mixture distributions are widely used due to their favorable properties.
They can approximate any continuous probability distribution arbitrarily well for sufficiently large \( M \) \citep{delon2020wasserstein}, and as they are linear combinations of Gaussian components, they retain many analytical properties of Gaussian distributions, including closed-form expressions for moments \citep{bishop2006pattern}.

In this work, we distinguish between \emph{homogeneous} and \emph{heterogeneous} Gaussian mixtures, as the former admit particularly efficient quantization procedures.
A Gaussian mixture is called homogeneous if all component covariance matrices are simultaneously diagonalizable, i.e., if there exists an orthogonal matrix $\mV\in \realNum^{d\times d}$ whose columns form a common eigenbasis such that $\mV^T\Sigma_i\mV$ is diagonal for all $i$. 
Otherwise, it is called heterogeneous. 
Despite this structural constraint, homogeneous mixtures form a dense family in the space of continuous probability distributions and can approximate any target distribution arbitrarily well \cite[Proposition~4.1]{nestoridis2011universal}.

\subsection{Quantization of Probability Distributions}
We next formalize the notion of quantization, which provides a mapping from continuous probability distributions, such as Gaussian mixtures, to discrete ones supported on a finite set of points.
For \(\sX\subseteq\realNum^n\), let \(\sC=\{\vc_i\}_{i=1}^N\) be a set of $N$ points in \(\sX\) and \(\bsR = \{\sR_i\}_{i=1}^N\) a partition of \(\sX\) into $N$ regions. 
The quantization operator \(\signature_{\bsR,\sC}:\realNum^n\rightarrow\realNum^n\) is defined as
\begin{equation}
    \signature_{\bsR,\sC}(\vx) = \sum_{i=1}^N\vc_i\indicator_{\sR_i}(\vx).
\end{equation}
Intuitively, \(\signature_{\bsR, \sC}\) maps every point $\vx\in\sR_i$ to the representative location $\vc_i$. 
Consequently, for any probability distribution $\Prob \in \sP(\sX)$, 
\begin{equation}
    \signature_{\bsR,\sC} \# \Prob = \sum_{i=1}^N \Prob(\sR_i) \delta_{c_i} \in \discProb_N(\sX).
\end{equation}
Here, $\signature_{\bsR,\sC} \# \Prob$ is called the \emph{quantization} of $\Prob$, effectively converting a possibly continuous distribution into a discrete one with support of size $N$ based on the specified locations. 
An example of the quantization operator is shown in Figure~\ref{fig:quantization_operator}.

\begin{figure*}[t]
    \centering
    \begin{subfigure}[b]{0.48\textwidth}
        \centering
        \includegraphics[width=\linewidth]{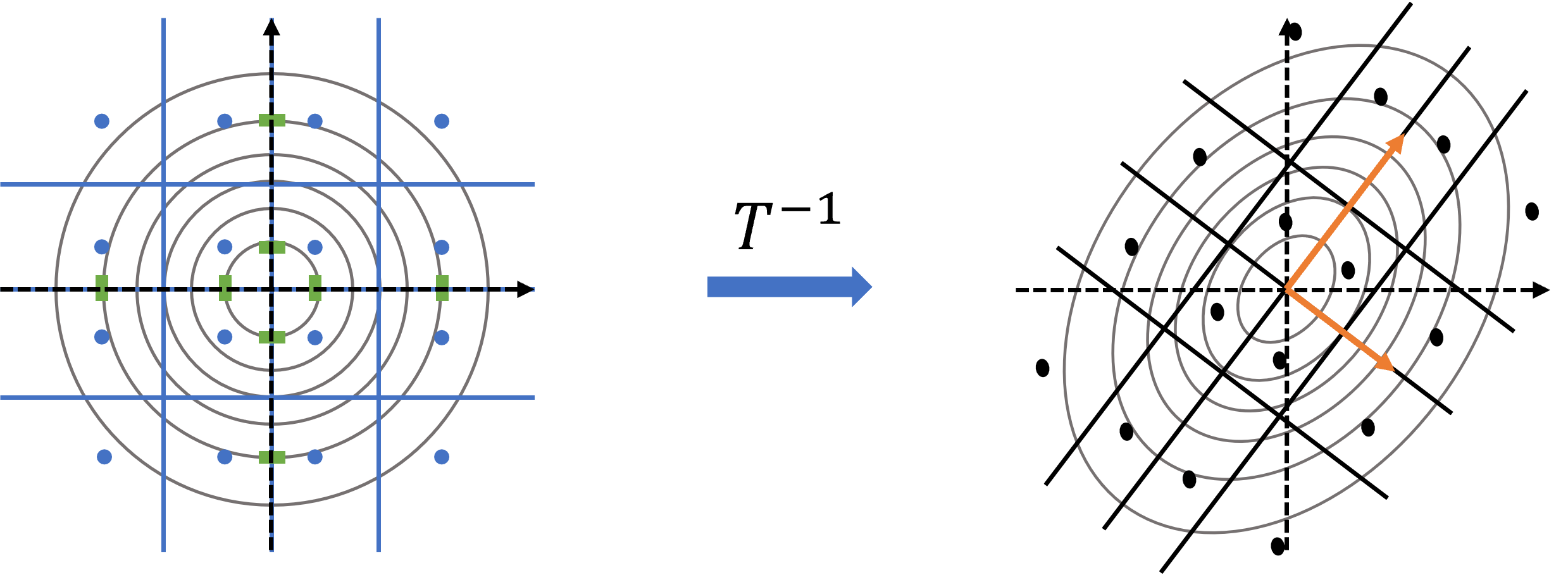}
        \caption{Grid-Configuration}\label{fig:grid}
    \end{subfigure}
    \hfill
    \begin{subfigure}[b]{0.48\textwidth}
        \centering
        \includegraphics[width=\linewidth]{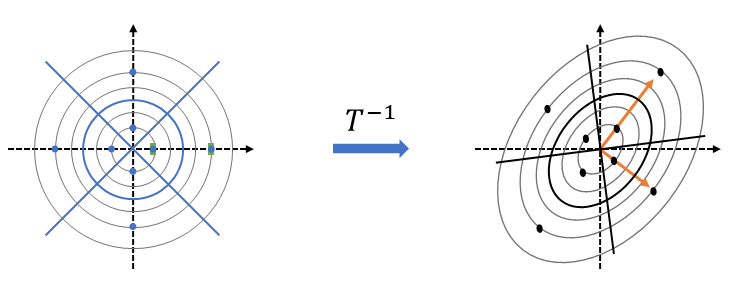}
        \caption{Cross-Configuration}\label{fig:cross}
    \end{subfigure}

    \caption{
    Quantization of a 2D Gaussian distribution (black dots) using a standard Gaussian (blue dots) as a template for grid and cross configurations. 
    In the eigenbasis \(T\) of the covariance matrix (orange arrows), the Gaussian’s dimensions are independent, allowing efficient construction of the discrete support. 
    In \ref{fig:grid}, the support is obtained by taking the cross-product of one-dimensional quantizations of the univariate marginals, whereas in \ref{fig:cross}, symmetric locations are placed along each principal axis (green markers). The resulting quantized distribution in the original space is obtained by mapping these locations through \(T^{-1}\). Blue and black lines indicate the Voronoi partitions in the transformed and original spaces, respectively.}
    \label{fig:grid_and_cross_quantization_gaussian}
\end{figure*}

The quantization \(\signature_{\bsR,\sC} \# \Prob\) can be computed particularly efficiently when the probability of each region $\Prob(\sR_i)$ admits a closed-form expression. 
In our tool, we exploit that for \(\Prob=\Ndist(\vmu, \Sigma)\), this occurs when the regions \(\sR_i\) are aligned with the geometry induced by the covariance matrix \(\Sigma\). 
In particular, tractable cases include: 
\begin{enumerate}[label=(\roman*)]
    \item \emph{Axis-aligned hyperrectangles} in the eigenbasis of \(\Sigma\), i.e., 
    \begin{equation}\label{eq:hyperrectangle}
        \sR_i = \{\vx\in\realNum^d \mid \va\leq\mT(\vx-\vmu)\leq\vb\},
    \end{equation}
    where $\Sigma=\mV\diag(\vlambda)\mV^T$ and $\mT=\diag(\vlambda)^{-1/2}\mV^T$, and vectors \(\va,\vb\in\realNum^d\).
    Specifically, for regions of this form, the probability of each region is given by
    \begin{equation}\label{eq:closed_form_hyperrectangles}
        \Prob(\sR_i) = \prod_{j=1}^d \left[\Phi\left(\evb{j}\right) - \Phi\left(\eva{j}\right)\right],
    \end{equation}
    where $\Phi$ denotes the standard normal cdf.
    \item \emph{Ellipsoids} parametrized by a threshold $\kappa>0$:
    \begin{equation}\label{eq:ellipsoid}
        \sR_i=\{\vx \in \realNum^d \mid (\vx - \vmu)^T \Sigma^{-1} (\vx - \vmu) \le \kappa\}.
    \end{equation}
    \item \emph{Unions or intersections} of such sets.
\end{enumerate}
Figure~\ref{fig:grid_and_cross_quantization_gaussian} illustrates examples of partitions constructed from regions of type (i) and (ii), each yielding closed-form quantizations for a Gaussian distribution.
Note that the tractability condition (i) naturally extends to homogeneous Gaussian mixtures, whose component covariance matrices share a common eigenbasis $\mV$. 

Lastly, note that the definition of $\signature_{\bsR,\sC}$ does not impose any relationship between the partition $\bsR$ and the locations $\sC$, allowing flexible, custom quantization schemes in our tool.
Nevertheless, in practice it is often natural to select $\bsR$ as the Voronoi partition of \(\sX\) induced by \(\sC=\{\vc_i\}_{i=1}^N\), where each region $\sR_i$ is defined by
\begin{equation}
    \sR_i = \left\{\vx\in\sX\mid \|\vx-\vc_i\|\leq\|\vx-\vc_j\|, \forall j\neq i \right\}.
\end{equation}
This partition minimizes the induced 2-Wasserstein quantization error in the Euclidean case, as discussed in the next subsection.

\subsection{Error in Wasserstein Distance}
To quantify the approximation error introduced by quantization, we use the \(2\)-Wasserstein distance, which measures the minimal transport cost required to transform one probability distribution into another. 
Formally, for \(\Prob, \ProbQ \in \sProb_2(\sX)\), it is defined as
\begin{equation}
    \wasserstein_2(\Prob, \ProbQ)
    = \left(
        \inf_{\gamma \in \Gamma(\Prob, \ProbQ)}
        \int_{\sX \times \sX}
        \|\vx - \vx'\|^2  \gamma(d\vx, d\vx')
      \right)^{1/2},
\end{equation}
where \(\Gamma(\Prob, \ProbQ) \subset \sProb(\sX \times \sX)\) denotes the set of all couplings (transportation plans) with marginals \(\Prob\) and \(\ProbQ\).

For our tool, we consider the natural coupling between \(\Prob\) and its quantization \(\signature_{\bsR,\sC} \# \Prob\), which transports all probability mass of \(\Prob\) contained in each region \(\sR_k\) to its representative location \(\vc_k\). 
This yields the upper bound 
\begin{equation}\label{eq:w2_error_quantization}
    \wasserstein_2(\Prob, \signature_{\bsR,\sC} \# \Prob)
    \leq
    \left(
        \sum_{k=1}^N
        \int_{\sR_k}
        \|\vx - \vc_k\|^2  d\Prob(\vx)
    \right)^{1/2}.
\end{equation}
which becomes an equality when \(\bsR\) is the Voronoi partition induced by \(\sC\)~\cite[Lemma~3.1]{canas2012learning}.
In this case, the constrained second moments in Eqn~\eqref{eq:w2_error_quantization} admit a closed-form expression for one-dimensional Gaussian distributions, as shown in \cite[Proposition 10]{adams2024finite}.

For homogeneous Gaussian mixtures, the quantization error separates into independent one-dimensional terms along the common eigenbasis of the component covariances, enabling closed-form and efficient computation of the Wasserstein distance \cite[Corollary~10]{adams2024finite}.
Specifically, for a Gaussian distribution \(\Ndist(\vmu,\Sigma) \allowbreak \in \GMM(\eucl^n)\) and a set of locations \(\sC\) forming an axis-aligned hyper-lattice in the space induced by a diagonalizing matrix $\mV$ of \(\Sigma\), with \(\bsR\) the Voronoi partition w.r.t. \(\sC\), the squared \(2\)-Wasserstein distance satisfies
\begin{equation}\label{eq:Was4SignMultGaus}
\begin{aligned}
    &\wasserstein_2^2\left(\signature_{\bsR,\sC}\#\Ndist(\vmu,\Sigma), \Ndist(\vmu,\Sigma)\right)
    = \\
    &\qquad\qquad\sum_{j=1}^n 
    \evlambda{j}
    \wasserstein_2^2\left(
        \signature_{\bsR_j, \sC_j}\#\Ndist(0,1), \Ndist(0,1)
    \right),
\end{aligned}
\end{equation}
where \(\evlambda{j}\) denotes the \(j\)-th eigenvalue of \(\Sigma=\mV\diag(\vlambda)\mV^T\), \(\sC_j\) is the set of quantization locations along the \(j\)-th dimension in the transformed (eigenbasis) space, and \(\bsR_j\) the corresponding one-dimensional Voronoi partition induced by \(\sC_j\).
The decomposition in Eqn~\eqref{eq:Was4SignMultGaus} extends directly to homogeneous Gaussian mixtures whose component covariances share a diagonalizing matrix \(\mV\), since
\begin{equation}\label{eq:error_mixtures}
    \wasserstein_2^2\left(\signature_{\bsR,\sC} \# \sum_{i=1}^M \elem{\vpi}{i} \Prob_i, \sum_{i=1}^M \elem{\vpi}{i} \Prob_i\right)
    = \sum_{i=1}^M
    \wasserstein_2^2(\signature_{\bsR,\sC} \# \Prob_i, \Prob_i).
\end{equation}

Unfortunately, the locations~$\sC$ that minimize the Wasserstein error for a given distribution are in general non-unique and computationally intractable even for multivariate Gaussians \cite{graf2000foundations}. An important exception is the univariate standard Gaussian, for which the optimal quantization locations can be computed efficiently~\cite{pages2003optimal} as utilized in our tool and discussed in the next section. 

\section{Efficient Quantization of Gaussian Mixtures}\label{sec:quantization_gmms}
We now present the algorithmic procedures implemented in our tool for quantizing Gaussian mixtures.
Following the tool's two-stage design, we first construct the tuples $(\bsR, \sC)$ of regions and locations, referred to as schemes, defining the quantization, and then apply the operation that produces the discrete approximation. 
While users can define custom schemes, the main functionality lies in the automatic generation of schemes that minimize the quantization error for a given distribution.
This automatic scheme-construction builds on modules for Gaussian distributions, which we first discuss before extending the procedure to Gaussian mixtures.

\subsection{Scheme Construction for Gaussians}
The tool supports the automatic generation of two classes of schemes for Gaussians that admit closed-form quantizations: grid-based schemes, corresponding to regions of type (i), and cross-based schemes, resembling sigma-based methods and corresponding to partitions of types (ii)–(iii), as illustrated in Figure \ref{fig:grid_and_cross_quantization_gaussian}.
In what follows, we focus on the former grid-based class, which provides formal Wasserstein guarantees and constitutes the main novelty of this tool.
The sigma-point procedure is well established in the literature \cite{julier2004unscented}, and its implementation within our framework is documented in detail in the accompanying \texttt{README}.

\begin{algorithm}[htbp]
\caption{Grid-scheme generation for Gaussian distributions}\label{alg:generate_scheme_gaussian}
\begin{algorithmic}[1]
    \Require $\Prob = \Ndist(\vmu, \Sigma)$ with $\Sigma = \mV \diag(\vlambda)\mV^\top$; support size $N$
    \State Lookup candidate grid layouts of size $N$
    \State Lookup optimal one-dimensional locations for each layout
    \State Compute $\wasserstein_2$ error per layout via Eqn~\eqref{eq:Was4SignMultGaus} using $\vlambda$
    \State \Return Optimal configuration and transformation matrix $\mT = \diag(\vlambda)^{-1/2}\mV^\top$ defining a non-axis-aligned grid
\end{algorithmic}
\end{algorithm}

Consider a multivariate Gaussian distribution \(\Ndist(\vmu,\Sigma)\) and an axis-aligned grid of regions as in Eqn~\eqref{eq:hyperrectangle}, defined in the space induced by the diagonalizing matrix \(\mV\) of \(\Sigma\). 
In this space, the distribution becomes standard normal and the grid becomes a Cartesian product of intervals, enabling the closed-form quantization of Eqn~\eqref{eq:closed_form_hyperrectangles}.
Similarly, the Wasserstein error in Eqn~\eqref{eq:w2_error_quantization} simplifies to a weighted sum of the one-dimensional errors with weights given by the eigenvalues of $\Sigma$. 
Consequently, the optimal grid assigns support points per dimension proportionally to these eigenvalues and selects the corresponding optimal one-dimensional partitions.

This yields a highly efficient quantization procedure, summarized in Algorithm~\ref{alg:generate_scheme_gaussian} and illustrated in Figure~\ref{fig:grid_and_cross_quantization_gaussian}.
Offline, two lookup tables are constructed: one storing grid layouts (used in line~1), which specify the possible combinations of grid points across dimensions for a given total grid size,
and one storing the optimal one-dimensional quantization locations for standard Gaussian distributions (used in line~2), computed efficiently following \cite{pages2003optimal}.
At run time, given a desired total support size and eigenbasis, the algorithm selects the configuration minimizing the Wasserstein error (lines 1–3) and stores it together with the transformation matrix parameterizing the non–axis-aligned grid (line 4).

\subsection{Scheme Generation for Gaussian Mixtures}\label{subsec:scheme_generation}
\begin{figure}[htbp]
    \centering
    \begin{subfigure}[b]{0.48\linewidth}
        \centering
        \includegraphics[width=\linewidth]{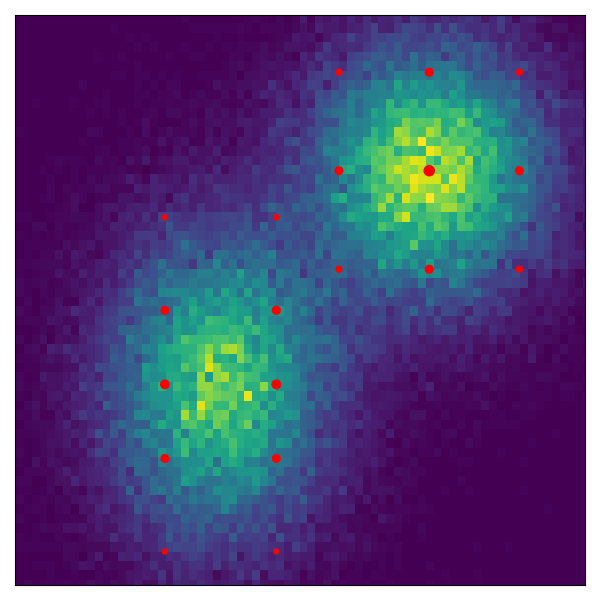}
        \caption{Per-Mode}\label{fig:per_mode}
    \end{subfigure}
    \hfill
    \begin{subfigure}[b]{0.48\linewidth}
        \centering
        \includegraphics[width=\linewidth]{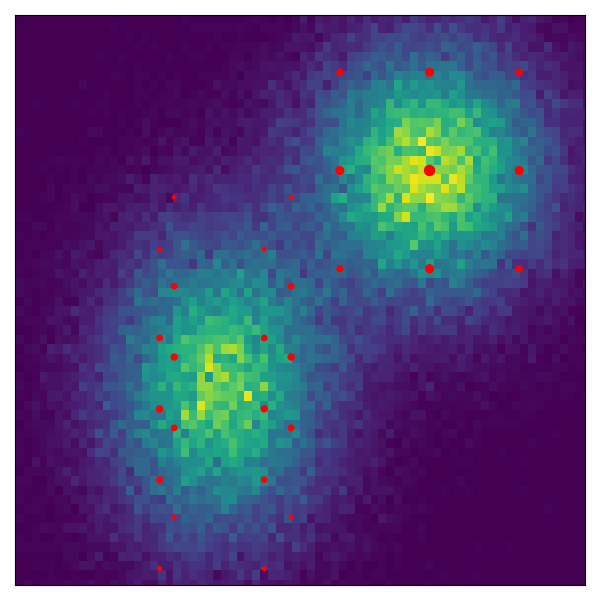}
        \caption{Per-Component}\label{fig:per_comp}
    \end{subfigure}
    \caption{
    Example 2D Gaussian mixture ($M=3$) used in Section~\ref{sec:tool} and its quantization (red dots) constructed per mode~\ref{fig:per_mode} and per component~\ref{fig:per_comp}. The mode-wise quantization achieves a comparable Wasserstein error ($0.47$ vs. $0.46$) while requiring a much smaller support ($19$ vs. $29$  points).}
    \label{fig:quantized_gmm}
\end{figure}

For Gaussian mixtures, particularly heterogeneous ones, there generally does not exist a single scheme that admits a closed form quantization for all components. 
A natural choice, as adopted in \cite{adams2024finite}, is therefore to define a separate scheme for each component, apply the quantization component-wise, and combine the resulting discrete distributions.\footnote{An alternative approach would be to use a common set of quantization locations for all mixture components and compute the resulting discrete distribution via numerical integration. However, this becomes computationally intractable when the component covariances differ significantly.} Formally, for a Gaussian mixture distribution $\Prob=\sum_{i=1}^M\evpi{i}\Prob_i$, we define a scheme \((\bsR_i, \sC_i)\) for each component $i$, and take as the discrete approximation
\(
    \sum_{i=1}^M \evpi{i} \signature_{\bsR_i,\sC_i}\# \Prob_i.
\)
Although such a component-wise procedure admits formal convergence guarantees \cite[Corollary~15]{adams2024finite}, it may allocate quantization points inefficiently when components are close, as illustrated in Figure~\ref{fig:quantized_gmm}. In these cases, post-hoc compression is often applied to remove redundant support points.

To address this, our tool constructs mode-wise schemes, as summarized in Algorithm~\ref{alg:generate_scheme_mixture}. 
The mixture's modes are first identified, and components are assigned to them based on their mean location (line~1). 
For grid-based schemes, the tool then distinguishes between clusters containing homogeneous or heterogeneous subsets of components.
For homogeneous clusters, a single grid scheme is generated using Algorithm~\ref{alg:generate_scheme_gaussian}, based on a local Gaussian approximation centered at the mode (line~4) and with a support size proportional to the mode’s weight (line~5).
For heterogeneous clusters, a separate per-component scheme is generated, and the resulting schemes are stored per mode to enable efficient mode-wise compression after quantization.
For non-grid schemes, all clusters are handled using the heterogeneous procedure.

\begin{algorithm}[tbhp]
\caption{Scheme generation for Gaussian mixtures}\label{alg:generate_scheme_mixture}
\begin{algorithmic}[1]
\algtext*{EndIf}
\algtext*{EndFor}
    \Require Gaussian mixture $\Prob = \sum_{i=1}^M \pi_i \, \Prob_i$, total support size $N$, Gaussian-scheme constructor $\textsc{SchemeGaussian}$ 
    \State Cluster components by mode based on mean locations
    \For{each cluster $\sS$}
        \State Allocate $N_{\sS}$ proportional to the cluster’s mixture weight
        \If{$\textsc{SchemeGaussian}=\textsc{Grid}$ \textbf{and} $\sS$ is homogeneous}
            \State Compute local Gaussian approximation at the mode 
            \State $\text{Scheme}_{\sS} \leftarrow \textsc{SchemeGaussian}(\text{local Gaussian}, N_{\sS})$
        \Else
            \For{each component $\Prob_i \in \sS$}
                \State Allocate $N_i$ proportionally to $\pi_i$ 
                \State $\text{Scheme}_i \leftarrow \textsc{SchemeGaussian}(\Prob_i, N_i)$
            \EndFor
            \State $\text{Scheme}_{\sS} \leftarrow \{\text{Scheme}_i : \Prob_i \in \sS\}$
        \EndIf
    \EndFor
    \State \Return $\{\text{Scheme}_{\sS}\}_{\text{modes}}$ \Comment{one scheme (or set) per mode}
\end{algorithmic}
\end{algorithm}

\subsection{Applying the Quantization Scheme}\label{subsec:quantization}
With the schemes defined, the final step applies them to obtain a discrete approximation.
The quantization operation accepts a Gaussian or Gaussian mixture and a user-defined or automatically generated scheme, and returns the corresponding discrete approximation together with a Wasserstein error bound when computable.

For a single scheme, the operation first verifies that the scheme’s reference frame shares the eigenbasis of the distribution’s covariance, i.e., that the regions satisfy constraints (i)–(iii).
The discretization is then computed in closed form; for grid-based schemes, its Wasserstein error is evaluated via Eqn~\eqref{eq:Was4SignMultGaus}.

For mixtures, the procedure generalizes based on the structure of the provided schemes.
If a single scheme is supplied, all components are quantized jointly after confirming alignment of their covariances with the scheme’s reference frame.
If multiple schemes are provided, the components are clustered by their mean locations, and each cluster is discretized using the scheme whose reference frame is closest to the cluster mean.
When a cluster contains a collection of component-specific schemes, the quantization is performed per component, followed by compression of redundant support points within each cluster using a weighted $k$-means procedure \cite{lloyd1982least}.

When the Wasserstein distance is tractable per scheme, the component-wise application of the quantization operation yields a computable upper bound on the total approximation error. 
Specifically, for $\Prob=\sum_{i=1}^M\evpi{i}\Prob_i$ denoting a mixture of homogeneous Gaussian components, and $\signature_{\bsR_i,\sC_i}\#\Prob_i$ the corresponding quantizations, it follows that 
\begin{equation}
    \wasserstein_2^2\left(\sum_{i=1}^M\signature_{\bsR_i,\sC_i} \#  \elem{\vpi}{i} \Prob_i, \sum_{i=1}^M \elem{\vpi}{i} \Prob_i\right)
    \leq \sum_{i=1}^M
    \wasserstein_2^2(\signature_{\bsR_i,\sC_i} \# \Prob_i, \Prob_i).
\end{equation}
Equality holds for $M=1$, i.e., the homogeneous case in Eqn~\eqref{eq:error_mixtures}.

\section{Tool Overview}\label{sec:tool}
\texttt{discretize\_distributions} is a modular PyTorch-based Python package implementing the quantization procedures from Section~\ref{sec:theory} for Gaussian mixture distributions.
The package provides three main components: the \texttt{.distributions} submodule for defining distribution classes, and two core functions, \texttt{generate\_scheme} for constructing quantization schemes, and \allowbreak\texttt{discretize}, which applies them.
In this section, we demonstrate how to construct distributions and quantization schemes, both automatically generated and custom, and how to apply the quantization operation. The source code is available at
\ifdefined\isarxiv
    \url{https://github.com/sjladams/discretize_distributions}.
\else
    \url{https://anonymous.4open.science/r/discretize_distributions-95F7/}.
\fi

\subsection{\texttt{.distributions} submodule}
The \texttt{.distributions} submodule provides the probabilistic building blocks of the package.
It extends PyTorch’s native distribution classes to support both the continuous input distributions to be quantized and the resulting discrete distributions.
The class \texttt{MultivariateNormal} extends PyTorch’s implementation to handle singular covariance matrices, and adds analytical Hessians and Jacobians.
\texttt{MixtureMultivariateNormal} specializes PyTorch’s \allowbreak\texttt{MixtureSameFamily} for Gaussian mixtures, providing closed-form expressions for their moments.
The example below illustrates how to construct a 2D Gaussian mixture with three components:

\begin{lstlisting}
import discretize_distributions.distributions as dd_dists

locs = torch.tensor([[1.0, 1.0], [-1.1, -1.3], [-0.9, -0.8]])
variances = torch.tensor([[0.5, 0.6],[0.4, 0.8],[0.5, 0.8]])
covs = torch.diag_embed(variances)
weights = torch.tensor([0.5, 0.25, 0.25])

comp = dd_dists.MultivariateNormal(locs, covs)
mix = dd_dists.Categorical(weights)
gmm = dd_dists.MixtureMultivariateNormal(mix, comp)
\end{lstlisting}

The discrete distributions obtained after quantization are encoded by \texttt{CategoricalFloat}, which extends PyTorch’s \allowbreak\texttt{Categorical} distribution to continuous-valued supports.
In addition, we provide a memory-efficient encoding for grid-structured
supports via \texttt{CategoricalGrid}, which uses the grid objects introduced next. The example below shows how to define a two-point 2D
discrete distribution using \texttt{CategoricalFloat}:

\begin{lstlisting}
disc = dd_dists.CategoricalFloat(locs=torch.tensor([[0., 0.], [1., 1.]]), probs=torch.tensor([0.5, 0.5]))
\end{lstlisting}

\begin{table*}
    \centering
    \begin{tabular}{lllll | cccc | cccc }
     & & & & & \multicolumn{4}{c|}{Quantization Error [$\wasserstein_2$]} & \multicolumn{4}{c}{Computation Time [sec]} \\ \midrule
    & name & comp. & dims. & modes & 10 & 100 & 1,000 & 10,000 & 10 & 100 & 1,000 & 10,000 \\ \midrule
    (a) & Bimodal benchmark of \cite{xu2022accurate} & 2     & 2     & 2     & 0.467 & 0.185 & 0.061 & 0.020 & 0.86  & 0.84  & 0.85  & 0.88 \\
    (b) & Example 4.2 of \cite{runnalls2007kullback} & 4     & 2     & 4     & 0.995 & 0.367 & 0.127 & 0.045 & 0.90  & 0.87  & 0.88  & 0.85 \\
    (c) &1D Filtering \cite{assa2018wasserstein} & 10    & 1     & 5     & 0.145 & 0.049 & 0.028 & 0.027 & 2.63  & 0.93  & 0.94  & 0.95 \\
    (d) & Unc. Prop. Prior of \cite{terejanu2008uncertainty} & 2     & 1     & 2     & 0.302 & 0.097 & 0.082 & 0.082 & 0.84  & 0.86  & 0.83  & 0.81 \\
    (e) & Double-spiral \cite{adams2025formal} & 8     & 2     & 2     & 1.414e-2 & 0.181e-2 & 0.019e-2 & 0.015e-2 & 1.15  & 1.09  & 1.19  & 1.19 \\
    (f) & BNN layer & 1     & 256   & 1     & 0.325 & 0.211 & 0.147 & 0.101 & 2.66  & 2.60  & 2.62  & 2.80 \\
    (g) &Degenerative & 4     & 10     & 2     & 0.400 & 0.046 & 0.039 & 0.015 & 1.13  & 1.09  & 1.06  & 1.07 \\
    \bottomrule
    \end{tabular}
    \caption{Quantization error and computation time for grid-based schemes of increasing support size across Gaussian mixture benchmarks with varying numbers of components, dimensions, and modes.}
    \label{tab:results}
\end{table*}

\subsection{\texttt{generate\_scheme} function}
The function \texttt{generate\_scheme} constructs quantization schemes automatically using the procedure in Section~\ref{subsec:scheme_generation} and summarized in Algorithm~\ref{alg:generate_scheme_mixture}.
It takes as input either a \texttt{MixtureMultivariateNormal} or \texttt{MultivariateNormal} instance, the desired support size, and optional configuration settings, such as whether to generate schemes mode-wise and whether to use a grid- or cross-based configuration (see Figure~\ref{fig:grid_and_cross_quantization_gaussian}).
It outputs a \texttt{Scheme} object that efficiently encodes the tuple $(\bsR, \sC)$ of regions and locations defining the quantization.

The example below illustrates the construction of a mode-wise grid-based scheme with 20 support points for the mixture above:

\begin{lstlisting}
import discretize_distributions as dd

scheme = dd.generate_scheme(gmm, configuration="grid", per_mode=True, scheme_size=20)
\end{lstlisting}

Note that \texttt{generate\_scheme} is a high-level helper that automates the construction of quantization supports. The underlying objects remain fully modular, and schemes can also be created manually using the base \texttt{Scheme} classes, allowing users to define custom partitions or support structures while retaining compatibility with the package.
The example below shows a manually defined grid-based scheme. The reference-frame axes are first specified, followed by the construction
of a grid of support points within that frame:

\begin{lstlisting}
from discretize_distributions.schemes import Axes, Grid

rot_mat=torch.tensor([[0., 1.], [1., 0.]])
scales =torch.tensor([0.5, 1.3])
offset=torch.tensor([-0.2, 0.2])
points_per_dim = [[-0.6, 0., 0.6], [-0.5, 0., 0.5]]

grid = Grid(points_per_dim, Axes(rot_mat, scales, offset))
\end{lstlisting}

\subsection{\text{discretize} function}
The function \texttt{discretize} performs the quantization operation described in Section~\ref{subsec:quantization}.
It takes either a \texttt{MultivariateNormal} or \texttt{MixtureMultivariateNormal} instance, together with a \texttt{Scheme}, and returns a \texttt{CategoricalFloat} distribution representing the discretized approximation, together with a Wasserstein error certificate when available. The example below shows a typical usage:

\begin{lstlisting}
disc_gmm, w2 = dd.discretize(gmm, scheme)
\end{lstlisting}

Figure~\ref{fig:quantized_gmm} illustrates the resulting discretized distribution and the original Gaussian mixture, together with the quantization obtained when the scheme is generated with \texttt{per\_mode = False}.

\section{Experimental Evaluation}\label{sec:experiments}
To demonstrate the effectiveness of \texttt{discretize\_distributions}, we benchmark the tool on four representative Gaussian mixtures from the literature and applications. 
To the best of our knowledge, no other publicly available implementation provides quantization of Gaussian mixtures with certified Wasserstein guarantees. 
The benchmarks cover a broad range of settings, as summarized in Table~\ref{tab:results}:
(a–b) Gaussian mixture quantization and compression benchmarks from~\cite{xu2022accurate,runnalls2007kullback},
(c–e) mixtures arising in uncertainty propagation and nonlinear filtering for dynamical systems~\cite{assa2018wasserstein,terejanu2008uncertainty,adams2025formal},
(f) the output distribution of a trained Bayesian neural network layer,
and (g) a custom degenerate case designed to assess robustness under singular covariances.
All experiments were run on a laptop with an Intel Core i7-10610U CPU and 16~GB of RAM.

The results are summarized in Table~\ref{tab:results}.
Across all benchmarks, the quantization error exhibits the characteristic multi-scale decay predicted by optimal quantization theory: it decreases rapidly for small support sizes and transitions to a slower, asymptotic convergence rate of order $\mathcal{O}(\ln N / N^2)$ as the support size $N$ increases~\cite{pages2003optimal}.
Computation times remain nearly constant across all configurations, confirming the efficiency of the grid-based scheme generation and indicating that even large support grids are constructed at nearly equal speed.
The benefit of the mode-wise quantization is clearly visible in cases such as (c) and (e), where multiple components share only a few modes, resulting in fast error decay with smaller support sizes, as also illustrated in Figure~\ref{fig:quantized_gmm}.
Example (f) demonstrates that the method scales efficiently to high-dimensional settings, where optimal location allocation across dimensions enables both low quantization error and fast computation.
Finally, the singular mixture (g) confirms that the implementation remains robust under degenerate covariance structures, a property essential for practical applications where such behavior commonly arises from strongly correlated or partially deterministic states.

\section{Conclusion}
We presented a Python tool for quantizing Gaussian mixture distributions, yielding discrete approximations with certified Wasserstein error bounds. By separating scheme construction from the quantization operation, the implementation enables flexible reuse. Future work includes extending the framework to non-Gaussian distributions with closed-form quantization and leveraging its differentiable PyTorch backend for learning-based applications.


\bibliographystyle{ACM-Reference-Format}
\bibliography{bibliography.bib}

\end{document}